\pdfoutput=1

\documentclass[11pt]{article}

\usepackage{acl}
\usepackage{amssymb}
\usepackage[ruled,vlined,linesnumbered]{algorithm2e}

\SetKwInput{KwParam}{Parameters}

\usepackage{times}
\usepackage{latexsym}
\usepackage{graphicx}
\usepackage[ruled,vlined,linesnumbered]{algorithm2e}

\usepackage{subcaption}
\usepackage{amsmath}
\usepackage{cleveref}

\usepackage{booktabs}
\usepackage{multirow}
\usepackage{graphicx} 

\usepackage[T1]{fontenc}

\usepackage[utf8]{inputenc}

\usepackage{microtype}

\title{CAS: Conformalized Agentic Search via Adaptive
Retrieval and Policy Weighting}

\author{
Zixi Zhu\textsuperscript{1*},
Jiayuan Su\textsuperscript{3*},
Jian Zhang\textsuperscript{1},
Yu Lin\textsuperscript{1\textdagger},
Hongwei Wang\textsuperscript{12\textdagger}
\\
\textsuperscript{1}ZJU-UIUC Institute, Zhejiang University. 
\\
\textsuperscript{2}State Key Laboratory of CAD\&CG, Zhejiang University.
\\
\textsuperscript{3}Tencent Inc.
\\
\texttt{\{zixizhu, 12221038, hongweiwang\}@zju.edu.cn}
\\
\texttt{yulin@intl.zju.edu.cn}
\\
\texttt{matt.jiayuan.su@gmail.com}
}

\begin{document}
\maketitle
\begingroup
\renewcommand{\thefootnote}{\fnsymbol{footnote}}
\footnotetext[1]{Equal contribution.}
\footnotetext[2]{Corresponding authors.}
\endgroup
\begin{abstract}
Search Agents face a severe reliability crisis during reinforcement learning (RL) fine-tuning. Heuristic Top-K retrieval often causes critical evidence loss or noise inclusion, while overconfidence induced by progressive RL leads to hallucinated answers and redundant searches. To build highly reliable agents, we introduce Conformal Prediction (CP) and propose Conformalized Agentic Search (CAS). This framework establishes reliability guarantees on both the retrieval and training sides: on the retrieval side, an Adaptive Prediction Set (APS), a specific CP realization, translates statistical coverage into dynamic document truncation to construct prediction sets that are adaptive in size; on the training side, Adaptive Conformal Inference (ACI), a dynamic CP algorithm, dynamically constructs prediction sets with controllable coverage to quantify answer confidence, which is then used to penalize low-confidence trajectories within the Group Relative Policy Optimization (GRPO) objective, ensuring the model learns only from reliable ones. Experiments across single-hop and multi-hop QA datasets demonstrate that our framework significantly improves reasoning accuracy while drastically reducing redundant tool invocations, establishing a highly reliable and efficient agent paradigm. Our code is available at \url{https://github.com/S1llyBird/CAS}.
\end{abstract}

\begin{figure*}[htbp]
    \centering
    \includegraphics[width=\textwidth]{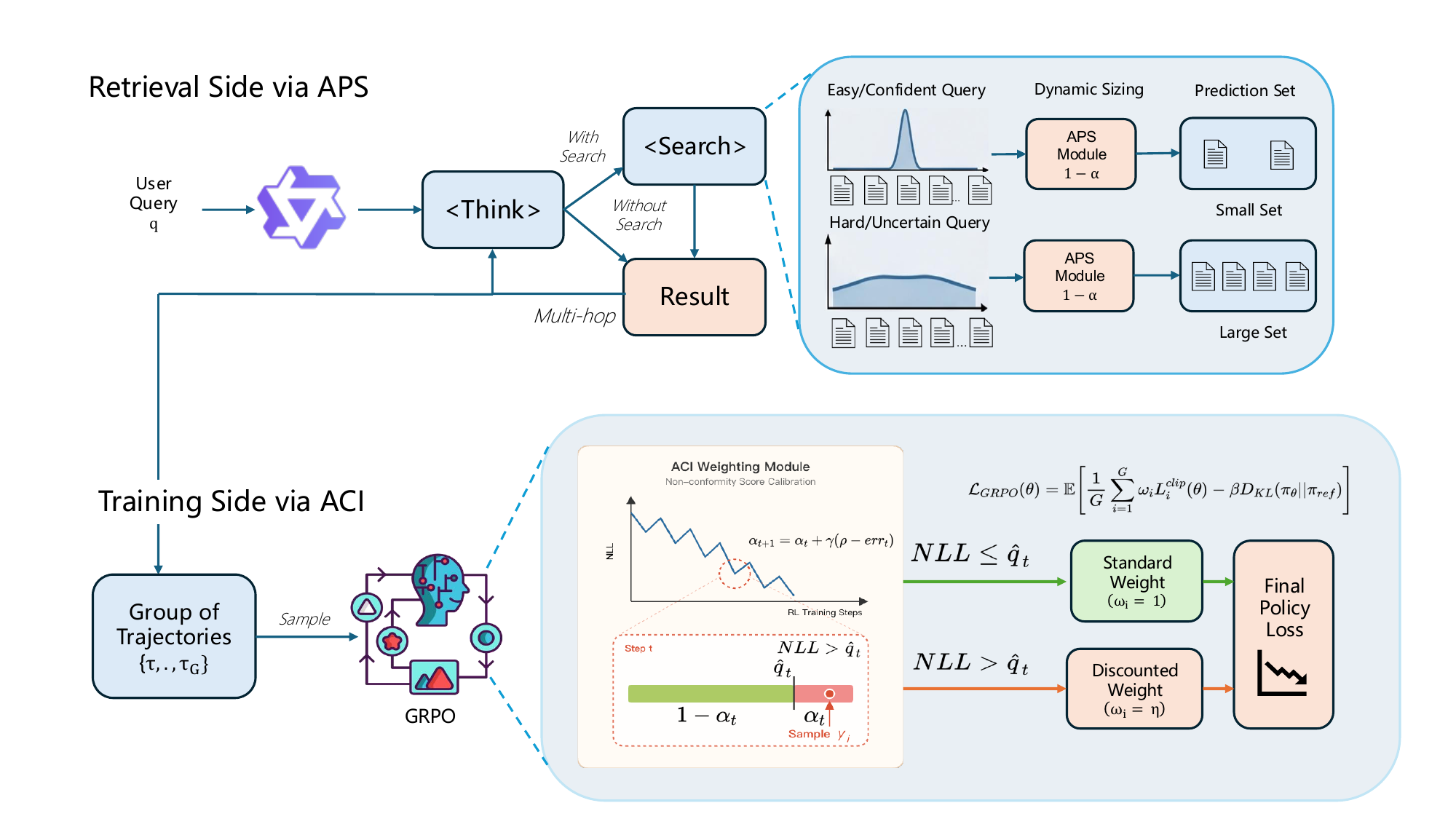} 
    
    \caption{\textbf{The CAS framework.} \textbf{(A)} APS dynamically sizes the retrieved document set based on local query difficulty. \textbf{(B)} ACI modulates the GRPO policy loss by penalizing low-confidence trajectories ($\mathrm{NLL} > q_{1-\alpha_t}$) to enforce reliable policy optimization.}
    \label{fig:architecture}
\end{figure*}

\section{Introduction}

Large Language Models (LLMs) have significantly advanced complex problem-solving by integrating external knowledge \cite{schick2023toolformerlanguagemodelsteach,liang2026voxmindendtoendagenticspoken}. While traditional Retrieval-Augmented Generation (RAG) employs a static "retrieve-then-generate" paradigm \cite{lewis2021retrievalaugmentedgenerationknowledgeintensivenlp}, the recent emergence of Agentic Search offers a more dynamic and autonomous approach \cite{jin2025searchr1trainingllmsreason,he2026searchr2enhancingsearchintegratedreasoning,zhao2025parallelsearchtrainllmsdecompose}. By interleaving internal reasoning steps with external information-gathering actions within a continuous generation trajectory, the agent autonomously plans when to retrieve and how to integrate newly acquired knowledge \cite{yao2023reactsynergizingreasoningacting}. 

However, fine-tuning these agents via reinforcement learning (RL) carries inherent risks of unreliability. First, on the retrieval side, heuristic Top-K truncation is inherently unreliable. Given varying query difficulties, a fixed K inevitably leads to the omission of critical facts or the inclusion of distracting noise \cite{liu2023lostmiddlelanguagemodels}. Second, on the training side, LLMs often exhibit overconfidence as RL progresses \cite{leng2025tamingoverconfidencellmsreward}, leading the model to generate hallucinated responses. Without effective confidence constraints, Search Agents are easily trapped in a cycle of highly inefficient, redundant tool invocations \cite{yin2026reasoningtrapenhancingllm}.

To address these risks, we adopt Conformal Prediction (CP) \cite{angelopoulos2022gentleintroductionconformalprediction}, a principled statistical framework that quantifies model uncertainty with rigorous theoretical guarantees. Unlike heuristic methods, CP provides strong finite-sample coverage guarantees: given a user-specified error rate $\alpha$, it constructs a prediction set that satisfies a target coverage level of at least $1-\alpha$. These properties make CP a theoretically grounded choice for establishing reliability in autonomous systems. Building on this, we propose Conformalized Agentic Search (CAS), which simultaneously applies CP to both the retrieval and training sides to provide rigorous statistical guarantees \cite{kumar2023conformalpredictionlargelanguage,su2025cprouteruncertaintyawarerouterllm}.  

On the retrieval side, we implement CP via the Adaptive Prediction Set (APS) method \cite{romano2020classificationvalidadaptivecoverage}. While ensuring a strict marginal coverage guarantee, APS dynamically adjusts the size of the set of retrieved items based on the model's "uncertainty" regarding the current input \cite{Chakraborty_2026}. On the training side, to mitigate the severe calibration degradation and overconfidence inherent in standard RL, we utilize Adaptive Conformal Inference (ACI) \cite{gibbs2021adaptiveconformalinferencedistribution}, a dynamic CP algorithm. ACI dynamically constructs prediction sets with controllable coverage to quantify answer reliability during training. We optimize the Group Relative Policy Optimization (GRPO) \cite{shao2024deepseekmathpushinglimitsmathematical} process by penalizing low-confidence trajectories (both blind overconfidence and erratic underconfidence), ensuring the model learns only from highly reliable reasoning paths.

In summary, our core contributions are threefold:

\begin{itemize}
\item A Reliable Theoretical Framework: We propose CAS, the first framework to introduce CP into the RL fine-tuning of search agents, providing rigorous statistical guarantees for reasoning and retrieval reliability.
\item CP Constraints on Both Sides: We implement APS on the retrieval side to construct prediction sets that are adaptive in size, ensuring the inclusion of correct evidence. Concurrently, we apply ACI on the training side to optimize the GRPO process, mitigating low-confidence outputs.
\item Superior Accuracy and Efficiency: Extensive experiments across single-hop and multi-hop QA datasets demonstrate that our framework significantly improves reasoning accuracy and training stability while drastically reducing redundant tool invocations, yielding a highly reliable and efficient agent paradigm.
\end{itemize}


\section{Conformal Prediction}

CP \cite{angelopoulos2022gentleintroductionconformalprediction} is a principled statistical framework that quantifies uncertainty with rigorous coverage guarantees, regardless of the underlying model or data distribution. Central to CP is a non-conformity score function $s(x,y)$, which measures the ``unusualness'' of a candidate output $y$ given an input $x$.

Let $(X, Y)$ be a sample, where $X$ represents the input and $Y$ represents the output. Suppose we have a calibration set of $n$ samples, denoted as $(X_i, Y_i)_{i=1}^n$, and a test sample $(X_{test}, Y_{test})$ drawn independently and identically (i.i.d.) from the same underlying distribution. Given a user-specified target error rate $\alpha \in (0,1)$, CP computes a quantile threshold $\hat{q}$ corresponding to the $\frac{\lceil(n+1)(1-\alpha)\rceil}{n}$ empirical quantile of the calibration scores. It then constructs a prediction set $\mathcal{C}_{1-\alpha}(X_{test})$ defined as:
\begin{equation}
\mathcal{C}_{1-\alpha}(X_{test}) = \{y \in \mathcal{Y} : s(X_{test},y) \le \hat{q}\}
\label{cp_general}
\end{equation}
Under the i.i.d. assumption, this procedure formally guarantees marginal coverage: $P(Y_{test} \in \mathcal{C}_{1-\alpha}(X_{test})) \ge 1 - \alpha$. See Appendix \ref{sec:appendix_cp_proof} for the formal proof.

\subsection{Adaptive Prediction Set}
\label{sec:3.1}

As a specific realization of the CP framework for classification and generative tasks, APS \cite{romano2020classificationvalidadaptivecoverage} constructs prediction sets by defining a specialized non-conformity score. Specifically, given sorted predictive probabilities $\pi_{(1)}(x)\ge\cdot\cdot\cdot\ge\pi_{(n)}(x)$, APS defines the non-conformity score $s(x,y)$ as the cumulative mass up to the true label $y$: $s=\sum_{j=1}^{L(y)}\pi_{(j)}(x)$, where $L(y)$ denotes the rank of $y$. During the CP inference phase, APS identifies the minimum index $k$ to form the prediction set such that the cumulative probability exceeds the calibrated threshold $\hat{q}$:
\begin{equation}
\sum_{i=1}^{k}\pi_{(i)}(x)\ge\hat{q}
\label{aps_general}
\end{equation}
This mechanism adaptively yields compact sets for confident inputs and expanded sets for ambiguous ones, strictly maintaining the $1-\alpha$ CP coverage guarantee while asymptotically approximating conditional coverage. See Appendix \ref{sec:appendix_aps_proof} for the formal proof and discussions on its conditional coverage.

\subsection{Adaptive Conformal Inference}
While the standard CP framework fundamentally relies on the exchangeability (i.i.d.) assumption, ACI \cite{gibbs2021adaptiveconformalinferencedistribution} is a dynamic extension designed to handle data streams where the underlying data distribution may change over time. Instead of maintaining a static target error rate, ACI introduces a time-varying error parameter $\alpha_{t}$.

At each time step $t$, we observe a test point $(X_{t}, Y_{t})$, where $X_{t}$ is the input and $Y_{t}$ is the true response. The algorithm evaluates whether $Y_{t}$ was contained within the previous prediction set via the empirical miscoverage indicator:
\begin{equation}
err_{t} := \begin{cases} 1, & \text{if } Y_{t} \notin \mathcal{C}_{t}(\alpha_{t}) \\ 0, & \text{otherwise} \end{cases}
\end{equation}
where $\mathcal{C}_{t}(\alpha_{t}) := \{y \in \mathcal{Y} : s(X_{t}, y) \le \hat{Q}_{t}(1-\alpha_{t})\}$ is the dynamic prediction set, and $\hat{Q}_{t}(\cdot)$ is the empirical quantile function.

Given a long-term target error rate $\rho$ and a step size $\gamma > 0$, ACI updates the error parameter via a simple online rule:
\begin{equation}
\alpha_{t+1} = \alpha_{t} + \gamma(\rho - err_{t})
\end{equation}
This recursive mechanism acts as a feedback loop: miscoverage ($err_{t}=1$) decreases $\alpha_{t}$, thereby expanding the subsequent prediction set to be more conservative. Conversely, success ($err_{t}=0$) increases $\alpha_{t}$, tightening the set. By continuously adapting $\alpha_{t}$, ACI preserves valid uncertainty quantification even when the data distribution changes over time. The formal proof of this dynamic guarantee is provided in Appendix \ref{sec:appendix_aci_proof}.







\section{Methodology}

We present the CAS framework (Figure~\ref{fig:architecture}) to reliably bridge LLMs' internal reasoning with external retrieval. We introduce two synergistic modules for statistical reliability: APS to bound dynamic retrieval uncertainty, and ACI to penalize low-confidence trajectories during policy optimization.

\subsection{Overview of Agentic Search}

Our policy model $\pi_\theta(y|x)$ employs a strict generative grammar to interleave internal reasoning with external actions. Given an input $x$, the model initiates reasoning within \texttt{<think>...</think>} tags.Upon reaching a knowledge boundary, it emits a query $q$ enclosed in \texttt{<search>...</search>} tags,, halting generation to invoke an external search engine $\mathcal{S}$. Crucially, rather than using a fixed top-$k$, the raw retrieved evidence $D = \mathcal{S}(q)$ is dynamically truncated into a reliable subset $D_{\text{APS}}$ via an APS. This subset is then wrapped in \texttt{<information>...</information>} tags and appended to the context. This generation-retrieval cycle repeats until the model outputs its final prediction $a_{\text{pred}}$ inside \texttt{<answer>...</answer>} tags. The complete prompt template is provided in Table \ref{tab:prompt_reasoning}.

\subsection{Retrieval Side via APS} \label{sec:aps}

Traditional tool-use frameworks typically append a fixed number of top-$k$ results to the context, which often injects redundant noise or truncates critical information. To rigorously bound the uncertainty of external evidence, we frame our dynamic retrieval mechanism within the general CP paradigm (Eq. \eqref{cp_general}). However, while standard CP methods guarantee marginal coverage across the data distribution, they fail to guarantee conditional coverage—often failing to adapt to the specific difficulty of a given input.

To heuristically approximate conditional coverage, we specify $\alpha_{APS}$ and construct a reliable document subset by implementing the APS detailed in Section \ref{sec:3.1}, thereby obtaining the calibrated threshold $\hat{q}_{APS}$. Given a query $q$, the search engine $\mathcal{S}$ returns an initial candidate set $D = \{d_1, d_2, \dots, d_n\}$ accompanied by their raw retrieval scores. We normalize these scores using a softmax operation to obtain a relevance probability $p(d_i|q)$ for each document. Following the rigorous APS inference procedure, our system identifies the truncation index $k$ by accumulating these probabilities until the mass first exceeds the calibrated threshold $\hat{q}_{\text{APS}}$. By dynamically adjusting to the conditional probability distribution of $q$, this process yields a statistically guaranteed subset $D_{\text{APS}} = \{d_1, \dots, d_k\}$. Crucially, this provides a statistical guarantee that ensures a $1-\alpha_{APS}$ coverage rate while adapting the context length to the difficulty of q.  

\subsection{Reward Design}

In our RL framework, we train the policy $\pi_{\theta}$ using a rule-based reward $r(x,y) = r_{acc} + r_{fmt}$. We define $r_{acc} = EM(a_{pred}, a_{gold}) \in \{0, 1\}$ as the exact match indicator. To enforce structural integrity, $r_{fmt}(y)$ incorporates two Boolean indicators, $\mathbb{I}_{\text{valid}}$ and $\mathbb{I}_{\text{ans}}$, representing strict grammatical correctness and the successful generation of the \texttt{<answer>} boundary, respectively. With a scaling factor $\gamma=0.2$, the format reward is formulated as: 
\begin{multline}
r_{\text{fmt}}(y) = \gamma \cdot \Big[ - r_{\text{acc}}(1 - \mathbb{I}_{\text{valid}}) \\
+ (1 - r_{\text{acc}}) \left( \mathbb{I}_{\text{valid}} + \frac{1}{2}\mathbb{I}_{\text{ans}}(1 - \mathbb{I}_{\text{valid}}) \right) \Big]
\end{multline}

This formulation rewards the correct format: when the answer is correct ($r_{\text{acc}}=1$), it strictly applies a $-\gamma$ penalty for format violations to prevent reward hacking. Conversely, when the answer is incorrect ($r_{\text{acc}}=0$), it provides a dense intermediate signal ($\gamma$ for full validity, or $\frac{1}{2}\gamma$ for partial structural effort) to guide the model toward using the correct format.

\subsection{Training Side via ACI}

Standard CP relies on the strict exchangeability (i.i.d.) assumption. However, during RL, the policy $\pi_{\theta}$ continuously evolves, and the prevalent use of binary rewards often leads to model overconfidence. To mitigate the redundant invocations and hallucinated outputs caused by this overconfidence, we employ ACI.

In our framework, at each RL iteration $t$, for the $i$-th sampled trajectory $y_{i}$ given input $x_t$, we define its non-conformity score $s(x_t, y_i)$ as the Negative Log-Likelihood (NLL) \cite{quach2024conformallanguagemodeling} of the generated tokens strictly within the \texttt{<answer>...</answer>} tags. Let $\mathcal{M}_{i}$ denote the set of token indices corresponding to these final answer tokens. The score is formulated as:
\begin{equation} \label{nll}
s(x_t, y_i)=\frac{\sum_{j\in\mathcal{M}_{i}}(-\log\pi_{\theta}(y_{j}|y_{<j},x_t))}{|\mathcal{M}_{i}|}
\end{equation}

Given a pre-specified target error rate $\rho$, instead of using the standard update rule, we employ a smoothed ACI update mechanism over previously observed data to mitigate local variations in the error rate \cite{gibbs2021adaptiveconformalinferencedistribution}. Specifically, we update the error rate $\alpha_{t}$ by evaluating the recent empirical miscoverage frequency using an exponentially weighted moving average of past errors:
\begin{equation} \label{newaci}
\alpha_{t+1} = \alpha_{t} + \gamma \left( \rho - \sum_{s=1}^{t} v_s err_s \right)
\end{equation}
where $\{v_s\}_{1\le s \le t}$ is a sequence of increasing weights such that $\sum_{s=1}^{t} v_s = 1$. In practice, we define the temporal weights with a smoothing factor of $0.95$ as 
$$v_s := \frac{0.95^{t-s}}{\sum_{s'=1}^{t} 0.95^{t-s'}}.$$
This approach effectively produces smoother trajectories for $\alpha_t$ with less local variation. The updated $\alpha_{t}$ is then used to dynamically adjust the quantile threshold $\hat{q}_t = \hat{Q}_t(1-\alpha_t)$. Specifically, the empirical quantile function $\hat{Q}_t(\cdot)$ is evaluated over a rolling calibration window of past scores from iteration $r = \max(1, t-2000)$ to $t-1$. This threshold $\hat{q}_t$ is subsequently employed to partition low-confidence trajectories.

We employ GRPO \cite{shao2024deepseekmathpushinglimitsmathematical} for the RL training. For a given input $x_t$ at iteration $t$, the policy samples a group of $G$ trajectories $\{y_1, y_2, \dots, y_G\}$. GRPO optimizes the policy by computing the relative advantage $A_i$ for each trajectory $y_i$, which is obtained by normalizing its reward $R_i$ within the group: $A_i = \frac{R_i - \text{mean}(\mathbf{R})}{\text{std}(\mathbf{R})}$. 

Concurrently, to penalize low-confidence sequences generated during training, we introduce a discount factor $\eta \in (0, 1)$ for samples falling into the low-confidence set. Accordingly, we minimize the ACI-guided GRPO loss function as follows:
\begin{equation} \label{grponew}
\resizebox{0.95\hsize}{!}{%
    $\mathcal{L}_{GRPO}(\theta)=\mathbb{E}\left[\frac{1}{G}\sum_{i=1}^{G}\omega_{i}L_{i}^{clip}(\theta)-\beta D_{KL}(\pi_{\theta}||\pi_{ref})\right]$%
}
\end{equation}
where $\beta D_{KL}$ is the KL divergence penalty against the reference model $\pi_{ref}$, and $L_{i}^{clip}(\theta)$ is the standard GRPO clipped objective function driven by the advantage $A_i$. The dynamic confidence-based weight $\omega_{i}$ is defined as:
\begin{equation} \label{w}
\omega_{i}=\begin{cases}1,& \text{if } s(x_t, y_i) \le \hat{q}_{t}\\ \eta,& \text{otherwise}\end{cases}
\end{equation}

Crucially, this weighting mechanism synergizes with the RL advantage $A_i$ to provide a dual-constraint on model reliability. For unconfident lucky guesses ($s(x_t, y_i) > \hat{q}_t$ with $A_i > 0$), the positive reinforcement is discounted by $\eta$, preventing the model from learning to guess. Conversely, if the model is confidently incorrect ($s(x_t, y_i) \le \hat{q}_t$ but yielding $A_i < 0$), the full weight ($\omega_i = 1$) ensures the model receives the maximum penalty. 

To effectively penalize low-confidence samples during early training while preventing over-penalization of relatively high-confidence trajectories classified as low-confidence after the model converges, we set the discount factor to $\eta=0.5$. The complete procedure of CAS is formally presented in Algorithm \ref{alg:conformal_search}.

\section{Experiments}

\begin{table*}[t]
\centering
\caption{The main results on seven datasets. $\dagger/\star$ represents in-domain/out-of-domain datasets. The best and second best performances are set as bold and underlined, respectively.}
\label{tab:main_results}
\resizebox{\textwidth}{!}{
\begin{tabular}{lcccccccc}
\toprule
\multirow{2}{*}{\textbf{Methods}} & \multicolumn{3}{c}{\textbf{General QA}} & \multicolumn{4}{c}{\textbf{Multi-Hop QA}} & \multirow{2}{*}{\textbf{Average}} \\
\cmidrule(lr){2-4} \cmidrule(lr){5-8}
& \textbf{NQ$^\dagger$} & \textbf{TriviaQA$^\star$} & \textbf{PopQA$^\star$} & \textbf{HotpotQA$^\dagger$} & \textbf{2WikiMultiHopQA$^\star$} & \textbf{Musique$^\star$} & \textbf{Bamboogle$^\star$} & \\
\midrule
Direct Inference & 0.106 & 0.288 & 0.108 & 0.149 & 0.244 & 0.020 & 0.024 & 0.134 \\
CoT & 0.023 & 0.032 & 0.005 & 0.021 & 0.021 & 0.002 & 0.000 & 0.015 \\
RAG & 0.348 & 0.544 & 0.387 & 0.255 & 0.226 & 0.047 & 0.080 & 0.270 \\
IRCoT & 0.111 & 0.312 & 0.200 & 0.164 & 0.171 & 0.067 & 0.240 & 0.181 \\
Search-o1 & 0.238 & 0.472 & 0.262 & 0.221 & 0.218 & 0.054 & 0.320 & 0.255 \\
SFT & 0.249 & 0.292 & 0.104 & 0.186 & 0.248 & 0.044 & 0.112 & 0.176 \\
R1-base & 0.226 & 0.455 & 0.173 & 0.201 & 0.268 & 0.055 & 0.224 & 0.229 \\
R1-instruct & 0.210 & 0.449 & 0.171 & 0.208 & 0.275 & 0.060 & 0.192 & 0.224 \\
Rejection Sampling & 0.294 & 0.488 & 0.332 & 0.240 & 0.233 & 0.059 & 0.210 & 0.265 \\
\midrule
Search-R1 (Qwen2.5-3B-Instruct) & 0.397 & 0.565 & 0.391 & 0.331 & 0.310 & 0.124 & 0.232 & 0.336 \\
Search-R1 (Qwen3-8B) & 0.440 & 0.631 & 0.418 & 0.372 & 0.355 & 0.157 & 0.430 & 0.400 \\
Search-R2 (Qwen3-8B)\footnotemark & \underline{0.477} & \textbf{0.676} & \textbf{0.466} & \underline{0.412} & \underline{0.405} & \underline{0.172} & \underline{0.512} & \underline{0.446} \\
\midrule
\textbf{Ours (Qwen2.5-3B-Instruct)} & 0.441 & 0.607 & 0.447 & 0.407 & 0.385 & 0.169 & 0.352 & 0.401 \\
\textbf{Ours (Qwen3-8B)} & \textbf{0.490} & \underline{0.668} & \underline{0.465} & \textbf{0.463} & \textbf{0.431} & \textbf{0.213} & \textbf{0.520} & \textbf{0.464} \\
\bottomrule
\end{tabular}
}
\end{table*}

\begin{table}[htbp]
\centering
\caption{Performance improvements of our method compared to Search-R1 and Search-R2 on General QA (single-hop) and Multi-Hop QA. $\Delta$ denotes the absolute performance gain.}
\label{tab:improvements}
\resizebox{\columnwidth}{!}{
\begin{tabular}{lccc}
\toprule
\textbf{Methods} & \textbf{General QA} & \textbf{Multi-Hop QA} & \textbf{Overall} \\
\midrule
\multicolumn{4}{c}{\textit{Qwen2.5-3B-Instruct}} \\
\midrule
Search-R1 & 0.451 & 0.249 & 0.336 \\
\textbf{Ours} & \textbf{0.498} & \textbf{0.328} & \textbf{0.401} \\
\textit{Improvement ($\Delta$)} & \textit{+0.047} & \textit{\textbf{+0.079}} & \textit{+0.065} \\
\midrule
\midrule
\multicolumn{4}{c}{\textit{Qwen3-8B}} \\
\midrule
Search-R1 & 0.496 & 0.329 & 0.400 \\
Search-R2 & 0.540 & 0.375 & 0.446 \\
\textbf{Ours} & \textbf{0.541} & \textbf{0.407} & \textbf{0.464} \\
\textit{Improvement ($\Delta$) vs. Search-R1} & \textit{+0.045} & \textit{\textbf{+0.078}} & \textit{+0.064} \\
\textit{Improvement ($\Delta$) vs. Search-R2} & \textit{+0.001} & \textit{+0.032} & \textit{+0.018} \\
\bottomrule
\end{tabular}
}
\end{table}

\subsection{Experimental Setup}

\paragraph{Datasets.}
To comprehensively evaluate CAS, we conduct experiments on seven diverse open-domain Question Answering (QA) datasets covering both single-hop and complex reasoning capabilities: NQ \cite{kwiatkowski-etal-2019-natural}, TriviaQA \cite{joshi2017triviaqalargescaledistantly}, PopQA \cite{mallen2023trustlanguagemodelsinvestigating}, HotpotQA \cite{yang2018hotpotqadatasetdiverseexplainable}, 2WikiMultihopQA (2Wiki) \cite{ho2020constructingmultihopqadataset}, MuSiQue \cite{trivedi2022musiquemultihopquestionssinglehop}, and Bamboogle \cite{press2023measuringnarrowingcompositionalitygap}. For training our RL framework, we construct a mixed training corpus using the training splits of NQ and HotpotQA.

\paragraph{Baselines.}
We compare our method against a comprehensive suite of competitive baselines, which can be logically categorized into four groups: inference without retrieval, including Direct Inference and CoT reasoning \cite{wei2023chainofthoughtpromptingelicitsreasoning}; inference with retrieval, comprising standard RAG \cite{lewis2021retrievalaugmentedgenerationknowledgeintensivenlp}, IRCoT \cite{trivedi2023interleavingretrievalchainofthoughtreasoning}, and Search-o1 \cite{li2025searcho1agenticsearchenhancedlarge}; fine-tuning based methods, which involve Supervised Fine-Tuning (SFT) \cite{chung2022scalinginstructionfinetunedlanguagemodels}, RL-based fine-tuning without search (R1) \cite{Guo_2025}, and rejection sampling with a search engine \cite{ahn2024largelanguagemodelsmathematical}; and finally, our reference backbone, Search-R1 \cite{jin2025searchr1trainingllmsreason}, along with Search-R2 \cite{he2026searchr2enhancingsearchintegratedreasoning}.

\paragraph{Implementation Details.}
We initialize our policy model using Qwen2.5-3B \cite{qwen2025qwen25technicalreport} and Qwen3-8B \cite{yang2025qwen3technicalreport}. For the retrieval module, we utilize the dense retriever E5 \cite{wang2024textembeddingsweaklysupervisedcontrastive}, paired with the 2018 Wikipedia dump \cite{karpukhin2020densepassageretrievalopendomain} as the external knowledge base. During the GRPO training phase, we set the group rollout size to $G = 5$ and sample 512 prompts per training step. To prevent infinite generation, the maximum number of assistant-search interaction rounds is capped at 4. The learning rate is set to $1 \times 10^{-6}$. For the ACI module, the step size is empirically set to $\gamma = 0.005$. All models are evaluated using the Exact Match (EM) metric. We provide more details in Appendix \ref{sec:experimental_details}.

\paragraph{Conformal Settings.} 
We set $\alpha_{\text{APS}} = 0.2$ and the target error rate $\rho = 0.25$ for ACI. The ACI calibration set $\mathcal{C}$ comprises 150 randomly sampled queries from a mixture of NQ and HotpotQA training splits. For ACI, we evaluate the untrained backbone on $\mathcal{C}$ to compute the initial non-conformity scores (Eq. \eqref{nll}), establishing the base empirical quantile $q_{\alpha_0}$ to bootstrap the dynamic tracking process. For APS calibration, we construct a distinct calibration set $\mathcal{C}_{\text{APS}}$. We utilize DeepSeek-V3.2 \cite{deepseekai2025deepseekv32pushingfrontieropen} to decompose multi-hop queries from the original 150 samples. These decomposed sub-queries, alongside the original single-hop questions, form $\mathcal{C}_{\text{APS}}$, which totals 239 queries. Given the target $\alpha_{\text{APS}} = 0.2$, this sample size provides a statistical error margin of $\epsilon = 0.026$ for the actual coverage \cite{angelopoulos2022gentleintroductionconformalprediction}. DeepSeek-V3.2 then acts as the judge to locate the ground-truth documents within $\mathcal{C}_{\text{APS}}$, yielding the calibrated threshold $\hat{q}_{\text{APS}}$ as formulated in Section \ref{sec:aps}.

\subsection{Main Results}

Table \ref{tab:main_results} and Table \ref{tab:improvements} present the comprehensive evaluation results of our method against all baselines across the seven datasets. We summarize the key findings as follows:

CAS achieves superior performance across all evaluated settings. Specifically, on the Qwen3-8B backbone, our method achieves the highest overall average score of 0.464, outperforming the strong baseline Search-R2 (0.446) and substantially surpassing Search-R1 (0.400). A similar trend is observed on the Qwen2.5-3B-Instruct backbone, where our method achieves an average score of 0.401, improving upon Search-R1 by an absolute margin of +0.065. This consistent superiority across different model scales highlights the generalizability and robustness of CAS.
\footnotetext{As Search-R2 is closed-source, we are unable to evaluate it on Qwen2.5-3B-Instruct.}

Our approach demonstrates exceptional performance across both complex multi-step and straightforward single-hop scenarios. As shown in Table \ref{tab:improvements}, on Multi-Hop QA datasets, our method yields massive gains, outperforming Search-R1 by +0.079 (3B) and +0.078 (8B), and surpassing the highly optimized Search-R2 by +0.032. On General QA (single-hop) tasks, our models also achieve highly competitive accuracy, with the 8B backbone significantly outperforming Search-R1 (+0.045) and slightly edging out Search-R2 (+0.001). This demonstrates the improvements of our framework: APS provides a retrieval set with marginal coverage and approximate conditional coverage to prevent the model from encountering hallucinations due to excessive context in simple queries or missing answers in complex ones; meanwhile, ACI effectively ensures high-confidence model outputs, preventing hallucinations and redundant tool invocations. Further experimental analysis regarding the Qwen3-8B backbone is deferred to Appendix \ref{sec:appendix_dynamics}.

\begin{table}[htbp]
\centering
\caption{Ablation study on Qwen2.5-3B-Instruct. The table presents the unablated framework, component-wise ablations, and sensitivity analyses for $\rho$ and $\alpha_{\text{APS}}$.Full results across all individual datasets are detailed in Table \ref{tab:detailed_ablation}.}
\label{tab:ablation}
\resizebox{\columnwidth}{!}{
\begin{tabular}{lccc}
\toprule
\textbf{Methods} & \textbf{General QA} & \textbf{Multi-Hop QA} & \textbf{Overall} \\
\midrule
Ours (Default) & 0.498 & 0.328 & 0.401 \\
\midrule
-ACI & 0.481 & 0.311 & 0.384 \\
-APS & 0.490 & 0.313 & 0.389 \\
\midrule
$\rho = 0.1$ & 0.497 & 0.302 & 0.386 \\
$\rho = 0.4$ & 0.495 & 0.273 & 0.368 \\
\midrule
$\alpha_{\text{APS}} = 0.35$ & 0.476 & 0.294 & 0.372 \\
$\alpha_{\text{APS}} = 0.05$ & 0.504 & 0.234 & 0.350 \\
\bottomrule
\end{tabular}
}
\end{table}

\begin{figure*}[htbp]
    \centering
    \begin{subfigure}[b]{0.32\textwidth}
        \centering
        \includegraphics[width=\textwidth]{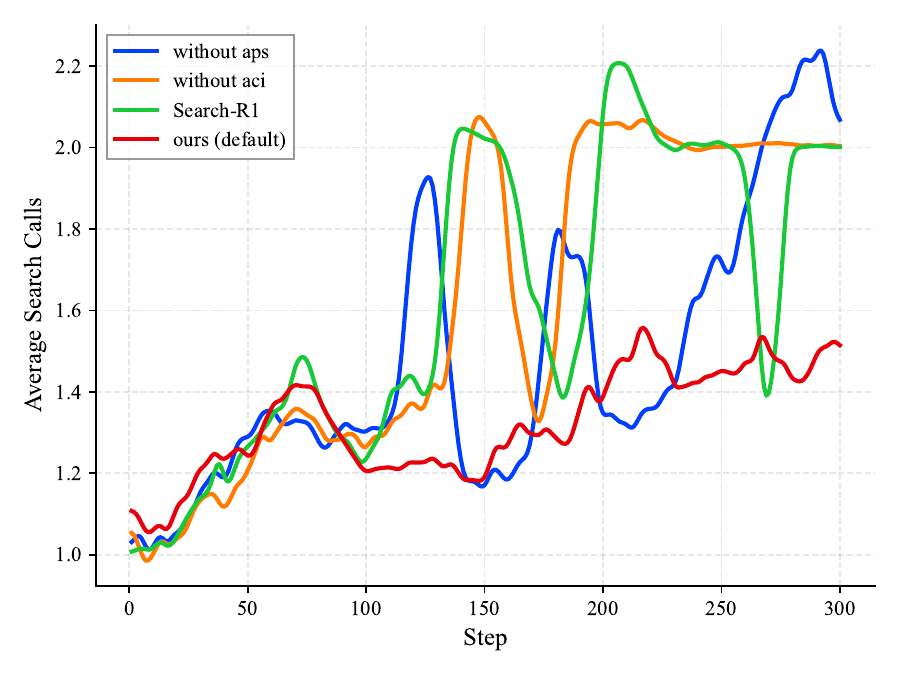}
        \caption{Average Search Calls}
        \label{fig:tool_calls}
    \end{subfigure}
    \hfill
    \begin{subfigure}[b]{0.32\textwidth}
        \centering
        \includegraphics[width=\textwidth]{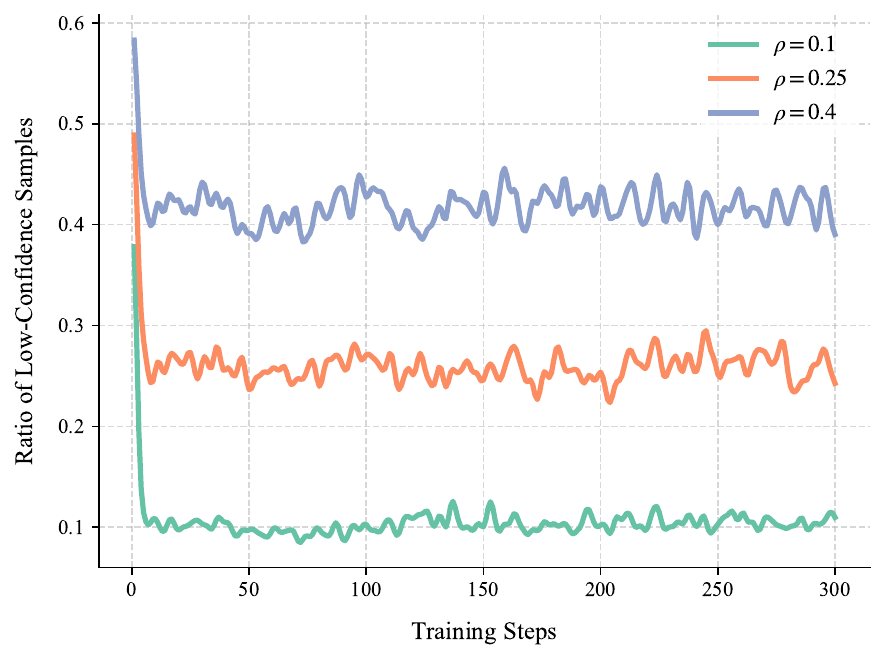}
        \caption{Ratio of Low-Confidence Samples}
        \label{fig:low_conf_ratio}
    \end{subfigure}
    \hfill
    \begin{subfigure}[b]{0.32\textwidth}
        \centering
        \includegraphics[width=\textwidth]{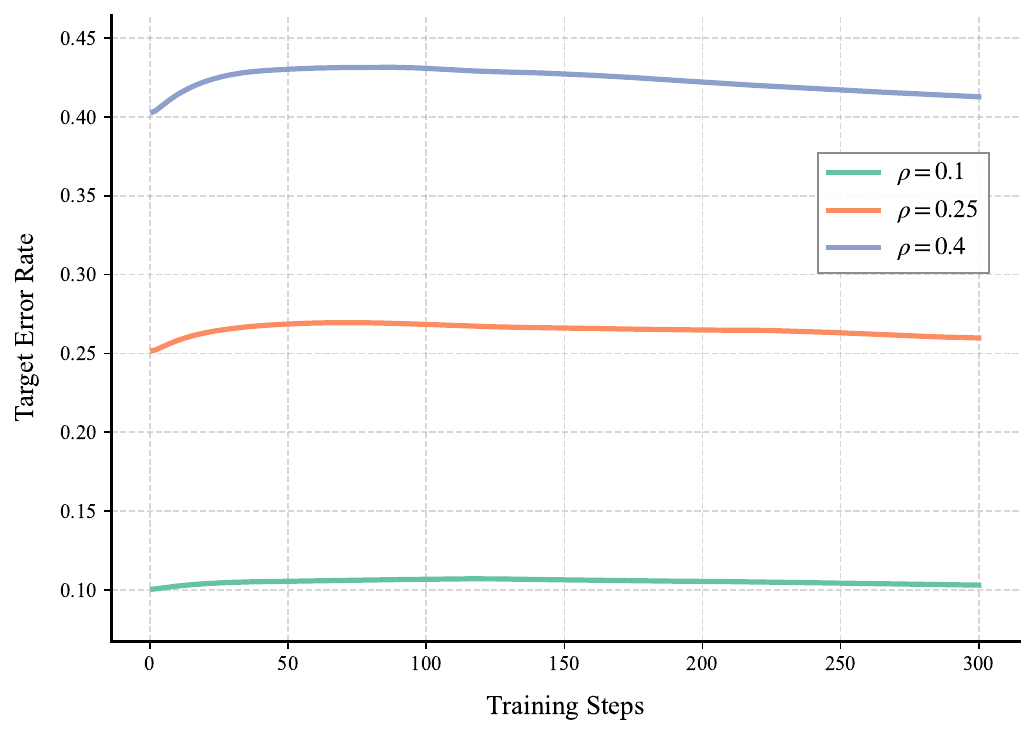}
        \caption{Adaptive Target Error Rate ($\alpha_t$)}
        \label{fig:alpha_evolution}
    \end{subfigure}
    
    \caption{Training dynamics and sensitivity analyses on the Qwen2.5-3B-Instruct backbone. \textbf{(a)} Evolution of average search calls during training across different ablation configurations. \textbf{(b)} The ratio of low-confidence samples penalized by the ACI weight under varying target error rates ($\rho$). \textbf{(c)} The dynamic adaptation of the target error rate ($\alpha_t$), demonstrating stable convergence to the preset $\rho$ values.}
    \label{fig:combined_dynamics}
\end{figure*}

\subsection{Ablation Study} \label{sec:ablation}
To evaluate the individual contributions of our proposed modules, we conducted a component-wise ablation study on the Qwen2.5-3B-Instruct backbone. For the configuration where the APS is disabled (-APS), the retrieval mechanism falls back to a fixed top-$k$ ($k=3$) setting. As shown in Table \ref{tab:ablation}, removing either component leads to a notable degradation in both General QA and Multi-Hop QA tasks.

\paragraph{Impact of the ACI Weight.}
Removing the ACI weight (-ACI) decreases the overall score from 0.401 to 0.384. This decline is intrinsically linked to the model's search behavior. As illustrated in the tool usage trajectories (Figure \ref{fig:combined_dynamics}(a)), the baseline and the -ACI variant exhibit significantly higher and more fluctuating tool calls. Without confidence constraints, blind overconfidence causes the model to hallucinate, initiating searches that deviate from the target question. Notably, some trajectories, due to a lack of confidence, conversely resort to secondary searches to verify answers. In summary, through the ACI constraint, the model avoids not only overconfidence but also blind underconfidence, thereby maintaining a stable and efficient search frequency (as denoted by the default trajectory).

\paragraph{Impact of the Adaptive Prediction Set.}
Disabling APS (-APS) drops accuracy to 0.389 by restricting retrieval to a fixed-length context. This rigid setup degrades performance via two paths: in simple queries, fixed top-$k$ retrieval introduces noise through redundant documents; in complex multi-hop queries, the static window often misses critical facts. Notably, the resulting information scarcity in complex scenarios forces the model to issue additional tool calls to compensate, even with the ACI weight active. Consequently, its tool usage frequency falls between our full framework and the baseline (Figure \ref{fig:tool_calls}). This dynamic corroborates the complementarity of the two modules: APS provides an adaptive, noise-free context in a single step, while the ACI weight suppresses unnecessary exploratory searches. 

\subsection{Sensitivity Analysis} \label{sec:sensitivity}
To verify the robustness and controllability of our framework, we conduct a sensitivity analysis on the ACI target error rate $\rho \in \{0.1, 0.25, 0.4\}$ and the APS significance level $\alpha_{\text{APS}} \in \{0.05, 0.2, 0.35\}$. All experiments in this section are performed on the Qwen2.5-3B-Instruct backbone, with calibration set configurations consistent with the Conformal Settings. The performance results are summarized in the bottom sections of Table \ref{tab:ablation}.

\paragraph{ACI Weight under Different $\rho$.} 
Figures \ref{fig:low_conf_ratio} and \ref{fig:alpha_evolution} illustrate the dynamic characteristics of the ACI mechanism during RL fine-tuning. The ACI weight effectively maintains the ratio of low-confidence samples within expected ranges and ensures that the dynamically adjusted $\alpha_t$ closely tracks the target error rate $\rho$. Notably, a pronounced spike is observed in the ratio of low-confidence samples at the very first step (figure \ref{fig:low_conf_ratio}). This phenomenon is directly attributed to the surge of trajectories as the policy $\pi_{\theta}$ begins to update, introducing highly non-i.i.d. data into the stream. The rapid stabilization of $\alpha_t$ following this shock demonstrates ACI's robust adaptability, highlighting the fundamental inadequacy of Static CP in dynamic RL environments. 

Furthermore, as shown in Table \ref{tab:ablation}, a strict target ($\rho = 0.4$) classifies nearly 40\% of the reasoning trajectories as low-confidence. While enforcing rigorous quality constraints, this over-penalization severely dilutes the RL reward signals, diminishing training efficiency. Conversely, a relaxed target ($\rho = 0.1$) applies the ACI weight to only 10\% of the samples. With such lenient filtering, the performance degenerates toward the unconstrained baseline due to insufficient confidence guidance.

\begin{table}[htbp]
\centering
\small
\caption{Average number of retrieved documents under different APS significance levels ($\alpha_{\text{APS}}$).}
\label{tab:aps_docs}
\begin{tabular}{lc}
\toprule
\textbf{$\alpha_{\text{APS}}$} & \textbf{Avg. Retrieved Documents} \\
\midrule
$0.20$ (Default) & 3.4 \\
$0.35$ & 2.4 \\
$0.05$ & 4.8 \\
\bottomrule
\end{tabular}
\end{table}

\paragraph{APS Retrieval under Different $\alpha_{\text{APS}}$.} 
The significance level $\alpha_{\text{APS}}$ dictates the aggressiveness of the dynamic context truncation. Table \ref{tab:aps_docs} presents the average number of retrieved documents under different $\alpha_{\text{APS}}$ settings. A high-guarantee setting ($\alpha_{\text{APS}} = 0.05$) yields an average of 4.8 documents, ensuring a 95\% marginal coverage. While this extensive context significantly benefits single-hop General QA by minimizing the risk of omitting critical evidence, the excessive information introduces substantial noise, which severely impairs the reasoning quality in complex Multi-Hop QA (see Table \ref{tab:ablation}). In contrast, a low-guarantee setting ($\alpha_{\text{APS}} = 0.35$) returns only 2.4 documents on average. This aggressive truncation fails to provide sufficient supporting facts, resulting in suboptimal performance across both tasks. Consequently, our default configuration ($\alpha_{\text{APS}} = 0.20$) strikes the optimal balance between comprehensive information retrieval and effective noise reduction.

\section{Related Works}

\subsection{Retrieval in LLMs}
Traditional RAG \cite{lewis2021retrievalaugmentedgenerationknowledgeintensivenlp,gao2024retrievalaugmentedgenerationlargelanguage} significantly expands the knowledge boundaries of LLMs by prepending retrieved external documents to the input context. With the continuous evolution of RAG, the emergence of frameworks such as Adaptive RAG \cite{jeong2024adaptiveraglearningadaptretrievalaugmented}, Search-o1 \cite{li2025searcho1agenticsearchenhancedlarge}, and SAKI-RAG \cite{tao-etal-2025-saki} has highlighted the inherent challenges of determining when to trigger retrieval in static paradigms. Concurrently, approaches that integrate retrieval with Reinforcement Learning \cite{jin2025searchr1trainingllmsreason,he2026searchr2enhancingsearchintegratedreasoning,zhao2025parallelsearchtrainllmsdecompose,singh2026agenticretrievalaugmentedgenerationsurvey}, have been introduced. However, these methods universally rely on fixed Top-K truncation. This static constraint fails to guarantee the marginal coverage of the retrieved knowledge.

\subsection{Reinforcement Learning for Agents}
Reinforcement Learning has fundamentally transformed the capabilities of LLMs, evolving them from passive generators into autonomous agents \cite{luo2025largelanguagemodelagent}, such as SWE-agents \cite{zhang2026toolenoughreinforcementlearning,wei2025swerladvancingllmreasoning}, Web Agents \cite{ding2026dynawebmodelbasedreinforcementlearning,guo2026opagentoperatoragentweb}, and Search Agents \cite{jin2025searchr1trainingllmsreason,zhao2025parallelsearchtrainllmsdecompose,he2026searchr2enhancingsearchintegratedreasoning,singh2026agenticretrievalaugmentedgenerationsurvey}. However, they universally face the problem of overconfidence driven by sparse, binary rewards, which subsequently leads to hallucination issues \cite{leng2025tamingoverconfidencellmsreward}. Recent work proposes training models to explicitly verbalize their confidence scores alongside their answers \cite{damani2025binaryrewardstraininglms}. Yet, this approach struggles in Search Agents, as the discontinuous generation process caused by continuous tool invocations prevents the model from explicitly expressing its confidence.

\subsection{Conformal Prediction and Uncertainty Quantification}
CP \cite{10.5555/1062391} offers highly reliable marginal coverage guarantees without distributional assumptions. Due to its theoretical rigor, CP has been widely applied in traditional classification and detection tasks \cite{wu2026filteringconfidencedataaugmentation, andéol2025conformalobjectdetectionsequential}. Recent works have integrated CP into LLMs  \cite{kumar2023conformalpredictionlargelanguage,quach2024conformallanguagemodeling,su2025cprouteruncertaintyawarerouterllm} and further into RL fine-tuning for robust alignment \cite{chen2026conformalfeedbackalignmentquantifying}. However, while CP guarantees marginal coverage, it fails to guarantee conditional coverage, exhibiting a lack of adaptability when faced with complex queries \cite{romano2020classificationvalidadaptivecoverage}. Furthermore, CP methods fundamentally rely on a strict independent and identically distributed (i.i.d.) assumption, which is inherently violated in reinforcement learning where the model and policy are continuously updating. Therefore, breaking these constraints is essential for reliable agentic search.

\section{Conclusion} \label{sec:conclusion}

We propose CAS, a framework that integrates Conformal Prediction to resolve the reliability crisis in RL-trained search agents. The synergy between APS and ACI ensures reliable document retrieval and mitigates model overconfidence. Empirically, CAS significantly enhances reasoning accuracy and reduces redundant tool invocations. By balancing theoretical rigor with practical performance, this work establishes a principled foundation for future reliable autonomous agents.

\section{Limitations}
\label{sec:limitations}

Although CAS demonstrates significant potential in improving the accuracy and efficiency of Search Agents, several limitations remain. First, our empirical validation is primarily focused on general open-domain Question Answering (QA) tasks. While CAS provides robust statistical guarantees within these general information-seeking contexts, its applicability in highly specialized professional domains remains unexplored. Second, the framework relies heavily on a strong external teacher model (e.g., DeepSeek-V3.2) to construct the calibration set for the APS by decomposing queries and acting as a relevance judge. Finally, CAS primarily focuses on outcome reliability without extending statistical guarantees to the intermediate reasoning process. Future research should explore verifiable process reliability.

\section*{Acknowledgments}

This work was supported by the National Key Research and Development Program of China (2024YFF0907802 and 2024YFF0907803) and the National Natural Science Foundation of China (62276230).

\bibliography{anthology,custom}

\clearpage
\appendix

\section{Pseudocode}
\label{sec:appendix1}

We provide the pseudocode of CAS in Algorithm \ref{alg:conformal_search}
\begin{algorithm}[htbp]
\caption{CAS}
\label{alg:conformal_search}
\SetAlgoLined
\small
\KwIn{Dataset $\mathcal{D}$, Initial policy $\pi_{\theta}$, Search engine $\mathcal{S}$}
\KwParam{Group size $G$, ACI step size $\gamma$, Target error rate $\rho$, Initial calibration set size $N_{\text{cal}}$}
\KwOut{Optimized policy $\pi_\theta$}
\BlankLine

Initialize ACI threshold $\alpha_1 \leftarrow \rho$\;
Initialize calibration set $\mathcal{C}$ with $N_{\text{cal}}$ scores from untrained policy\;

\For{each RL training iteration $t = 1, 2, \dots$}{
    Sample a prompt $x_t \sim \mathcal{D}$\;
    
    \For{trajectory $i = 1$ \KwTo $G$}{
        \While{trajectory $y_i$ not terminated}{
            Generate tokens $y_{next} \sim \pi_\theta(\cdot \mid x_t, y_i)$\;
            \If{search query $q$ generated}{
                $D_{\text{APS}} \leftarrow \text{APS}(\mathcal{S}(q))$ (Sec. \ref{sec:aps})\;
                $y_i \leftarrow y_i \oplus D_{\text{APS}}$\;
            }
        }
    }
    
    \For{trajectory $i = 1$ \KwTo $G$}{
        Compute reward $R_i = r_{\text{acc}} + r_{\text{fmt}}$\;
        Compute NLL score $s(x_t, y_i)$ for answer tokens (Eq. \eqref{nll})\;
    }
    Compute advantages $A_1, \dots, A_G$ from rewards $\{R_i\}$\;
    
    Compute quantile threshold $\hat{q}_t \leftarrow \hat{Q}_t(1-\alpha_t)$ over $\mathcal{C}$\;
    
    \For{trajectory $i = 1$ \KwTo $G$}{
        Determine ACI confidence weight $\omega_i$ using $\hat{q}_t$ (Eq. \eqref{w})\;
    }
    $\mathcal{C} \leftarrow \mathcal{C} \cup \{s(x_t, y_1), \dots, s(x_t, y_G)\}$ \;
    Update $\pi_\theta$ by minimizing the ACI-guided objective (Eq. \eqref{grponew})\;
    Update ACI threshold $\alpha_{t+1}$ (Eq. \eqref{newaci})\;
}
\end{algorithm}

\section{Mathematical Proofs}
\label{sec:appendix_proofs}

In this section, we provide the formal mathematical proofs for the statistical guarantees of the CP methods used in CAS. We begin with the fundamental marginal coverage guarantee of standard Split CP.

\subsection{Marginal Coverage Guarantee of Conformal Prediction}
\label{sec:appendix_cp_proof}

Following the standard proof of validity for split-conformal prediction \citep{angelopoulos2022gentleintroductionconformalprediction}, we demonstrate that the prediction sets constructed via CP possess a strict, finite-sample marginal coverage guarantee. 

\vspace{1.5mm}
\noindent \textbf{Theorem 1} \textit{(Conformal calibration coverage guarantee).} 
\textit{Suppose the calibration data $(X_i, Y_i)_{i=1,\dots,n}$ and the test point $(X_{\text{test}}, Y_{\text{test}})$ are independent and identically distributed (i.i.d.). Define the conformal quantile $\hat{q}$ as:}
{\small
\begin{equation}
    \hat{q} = \inf \left\{ q : \frac{|\{i : s(X_i, Y_i) \leq q\}|}{n} \geq \frac{\lceil(n+1)(1-\alpha)\rceil}{n} \right\}
\end{equation}
}
\textit{and the resulting prediction sets as:}
\begin{equation}
    \mathcal{C}(X) = \{y : s(X, y) \leq \hat{q}\}
\end{equation}
\textit{Then, the marginal coverage satisfies:}
\begin{equation}
    \mathbb{P}(Y_{\text{test}} \in \mathcal{C}(X_{\text{test}})) \geq 1 - \alpha
\end{equation}

\vspace{1.5mm}
\noindent \textit{Proof of Theorem 1.} 
Let $s_i = s(X_i, Y_i)$ for $i = 1, \dots, n$ and $s_{\text{test}} = s(X_{\text{test}}, Y_{\text{test}})$. To avoid handling ties, we consider the case where the non-conformity scores $s_i$ are distinct with probability 1. 

Without loss of generality, we assume the calibration scores are sorted such that $s_1 < s_2 < \dots < s_n$. In this case, the quantile $\hat{q}$ can be explicitly written as:
\begin{equation}
    \hat{q} = s_{\lceil(n+1)(1-\alpha)\rceil}
\end{equation}
when $\alpha \geq \frac{1}{n+1}$, and $\hat{q} = \infty$ otherwise. 

Note that in the case where $\hat{q} = \infty$, the prediction set includes the entire label space, i.e., $\mathcal{C}(X_{\text{test}}) = \mathcal{Y}$, so the coverage property is trivially satisfied. Thus, we only need to handle the case when $\alpha \geq \frac{1}{n+1}$.

We proceed by noticing the strict equality of the two following events:
\begin{equation}
    \{Y_{\text{test}} \in \mathcal{C}(X_{\text{test}})\} = \{s_{\text{test}} \leq \hat{q}\}
\end{equation}

Combining this with our definition of the sorted quantile $\hat{q}$ yields:
{\small
\begin{equation}
    \{Y_{\text{test}} \in \mathcal{C}(X_{\text{test}})\} = \{s_{\text{test}} \leq s_{\lceil(n+1)(1-\alpha)\rceil}\}
\end{equation}
}

Now comes the crucial insight: by the exchangeability of the random variables $(X_1, Y_1), \dots, (X_{\text{test}}, Y_{\text{test}})$, their corresponding non-conformity scores $s_1, \dots, s_n, s_{\text{test}}$ are also exchangeable. Because they are exchangeable, $s_{\text{test}}$ is equally likely to fall anywhere between the sorted calibration points $s_1, \dots, s_n$. Therefore, the probability that $s_{\text{test}}$ is less than or equal to the $k$-th sorted score is exactly:
\begin{equation}
    \mathbb{P}(s_{\text{test}} \leq s_k) = \frac{k}{n+1}
\end{equation}
for any integer $k$. (Note that here, the randomness is over all variables $s_1, \dots, s_n, s_{\text{test}}$).

From this property, we substitute $k = \lceil(n+1)(1-\alpha)\rceil$ to conclude:
{\small
\begin{equation}
    \mathbb{P}\left(s_{\text{test}} \leq s_{\lceil(n+1)(1-\alpha)\rceil}\right) = \frac{\lceil(n+1)(1-\alpha)\rceil}{n+1} \geq 1 - \alpha
\end{equation}
}
which implies the desired result: $\mathbb{P}(Y_{\text{test}} \in \mathcal{C}(X_{\text{test}})) \geq 1 - \alpha$.

\vspace{2mm}
\noindent \textbf{Theorem 2} \textit{(Conformal calibration upper bound).} \textit{Additionally, if the scores $s_1, \dots, s_n$ have a continuous joint distribution (i.e., avoiding ties), the coverage is tightly bounded from above:}
\begin{equation}
    \mathbb{P}(Y_{\text{test}} \in \mathcal{C}(X_{\text{test}})) \leq 1 - \alpha + \frac{1}{n+1}
\end{equation}
\textit{(Proof deferred to Theorem 2.2 of \citet{lei2017distributionfreepredictiveinferenceregression}).}

\subsection{Statistical Guarantees of Adaptive Prediction Sets}
\label{sec:appendix_aps_proof}

As established in the foundational literature, achieving exact finite-sample conditional coverage is theoretically impossible without strong distributional assumptions. However, the APS framework \cite{romano2020classificationvalidadaptivecoverage} effectively circumvents this limitation. It provides a rigorous marginal coverage guarantee while sensibly approximating conditional coverage by adapting the prediction set size to the local uncertainty of the input.

To construct the adaptive sets, APS introduces a generalized inverse quantile conformity score. Given a base model's probability estimate $\hat{\pi}$ and a uniform random variable $U \sim \text{Uniform}(0, 1)$ for tie-breaking, the conformity score function $E$ is defined as:
{\small
\begin{equation}
    E(x, y, u; \hat{\pi}) = \min \{\tau \in [0, 1] : y \in \mathcal{S}(x, u; \hat{\pi}, \tau)\}
\end{equation}
}
where $\mathcal{S}$ is the generalized conditional quantile function that includes classes in descending order of their estimated probabilities until the cumulative mass reaches $\tau$.

Using this conformity score, APS achieves the following rigorous marginal guarantee:

\vspace{1.5mm}
\noindent \textbf{Theorem 3} \textit{(Marginal coverage of APS).} 
\textit{If the calibration samples $(X_i, Y_i)_{i \in \mathcal{I}_2}$ and the test sample $(X_{\text{test}}, Y_{\text{test}})$ are exchangeable, and the conformity scores are calculated using a model trained on a disjoint split $\mathcal{I}_1$, the APS prediction set $\hat{\mathcal{C}}_{\text{APS}}$ satisfies:}
\begin{equation}
    \mathbb{P}\left(Y_{\text{test}} \in \hat{\mathcal{C}}_{\text{APS}}(X_{\text{test}})\right) \geq 1 - \alpha
\end{equation}
\textit{Furthermore, if the scores $E_i$ are almost surely distinct, the coverage is bounded tightly from above by $1 - \alpha + 1/(|\mathcal{I}_2| + 1)$.}

\vspace{1.5mm}
\noindent \textit{Proof of Theorem 3.} 
Let $E_i = E(X_i, Y_i, U_i; \hat{\pi})$ denote the conformity score for the $i$-th calibration sample in $\mathcal{I}_2$, and $E_{\text{test}} = E(X_{\text{test}}, Y_{\text{test}}, U_{\text{test}}; \hat{\pi})$ for the test point. 

By the construction of the APS prediction set, a label $y$ is included in $\hat{\mathcal{C}}_{\text{APS}}(X_{\text{test}})$ if and only if its requisite cumulative mass $\tau$ is less than or equal to the calibrated threshold $\hat{Q}_{1-\alpha}$. Mathematically, we know that:
{\small
\begin{equation}
    Y_{\text{test}} \in \hat{\mathcal{C}}_{\text{APS}}(X_{\text{test}}) \iff E_{\text{test}} \leq \hat{Q}_{1-\alpha}(\{E_i\}_{i \in \mathcal{I}_2})
\end{equation}
}
where $\hat{Q}_{1-\alpha}(\{E_i\}_{i \in \mathcal{I}_2})$ is defined as the $\lceil(1 - \alpha)(1 + |\mathcal{I}_2|)\rceil$-th smallest value in the calibration score set $\{E_i\}_{i \in \mathcal{I}_2}$.

Because the data points $(X, Y)$ are exchangeable and the uniform variables $U$ are i.i.d., all the evaluated conformity scores $E_{\text{test}}$ and $\{E_i\}_{i \in \mathcal{I}_2}$ are completely exchangeable. Under the property of exchangeability, the rank of $E_{\text{test}}$ is uniformly distributed among the $|\mathcal{I}_2| + 1$ scores. Therefore, the probability of the event that $E_{\text{test}}$ falls below the empirical $(1-\alpha)$-quantile is bounded from below by the nominal level:
\begin{equation}
    \mathbb{P}\left(E_{\text{test}} \leq \hat{Q}_{1-\alpha}(\{E_i\}_{i \in \mathcal{I}_2})\right) \geq 1 - \alpha
\end{equation}
which immediately establishes $\mathbb{P}(Y_{\text{test}} \in \hat{\mathcal{C}}_{\text{APS}}(X_{\text{test}})) \geq 1 - \alpha$. 

\vspace{1.5mm}
\noindent \textbf{Asymptotic Conditional Coverage.} 
While Theorem 3 guarantees marginal coverage, the structural design of APS provides an asymptotic approximation of conditional coverage. Consider an Oracle model with perfect knowledge of the true conditional distribution $\pi_y(x) = \mathbb{P}(Y=y | X=x)$. The Oracle's prediction set $\mathcal{C}_{\alpha}^{\text{oracle}}(x)$ naturally attains exact conditional coverage. 

According to \citet{romano2020classificationvalidadaptivecoverage}, as the sample size increases and if the base predictive model is consistent (i.e., $\hat{\pi}_y(x) \approx \pi_y(x)$), the constructed sets $\mathcal{S}(X, U; \hat{\pi}, \tau)$ will converge to contain the true labels for exactly a fraction $\tau$ of the points. In this limit, the threshold $\hat{Q}_{1-\alpha} \approx 1 - \alpha$, and the decision rule approaches:
{\small
\begin{equation}
    \hat{\mathcal{C}}_{\text{APS}}(X_{\text{test}}) \approx \{y \in \mathcal{Y} : E(X_{\text{test}}, y, U_{\text{test}}; \pi) \leq 1 - \alpha\}
\end{equation}
}
which mathematically equates to the exact output of the Oracle procedure, thereby closely approximating optimal conditional coverage in complex data scenarios.

\subsection{Statistical Guarantees of Adaptive Conformal Inference}
\label{sec:appendix_aci_proof}

Standard CP fundamentally relies on the exchangeability of the data. In online settings, the policy continuously evolves, leading to severe distribution shifts that violate the i.i.d. assumption. To maintain rigorous coverage, we employ ACI \cite{gibbs2021adaptiveconformalinferencedistribution}. 

ACI guarantees the target coverage frequency over long-time intervals irrespective of the true data-generating process by dynamically adjusting the nominal error level. Following \citet{gibbs2021adaptiveconformalinferencedistribution}, let $\rho \in (0, 1)$ be the target miscoverage rate. At each time step $t$, the algorithm uses a parameter $\alpha_t$ to construct the prediction set $\hat{\mathcal{C}}_t(\alpha_t)$, and records the miscoverage event:
\begin{equation}
    \text{err}_t := \begin{cases} 
    1, & \text{if } Y_t \notin \hat{\mathcal{C}}_t(\alpha_t) \\
    0, & \text{otherwise} 
    \end{cases}
\end{equation}
The parameter $\alpha_t$ is recursively updated using a step size $\gamma > 0$:
\begin{equation}
    \alpha_{t+1} := \alpha_t + \gamma(\rho - \text{err}_t)
\end{equation}

To establish the distribution-free guarantee, we assume that with probability one, $\alpha_1 \in [0, 1]$ and the quantile function $\hat{Q}_t(x)$ is non-decreasing with $\hat{Q}_t(x) = -\infty$ for $x < 0$ and $\hat{Q}_t(x) = \infty$ for $x > 1$.

\vspace{1.5mm}
\noindent \textbf{Lemma 4} \textit{(Boundedness of $\alpha_t$, Lemma 4.1 in \citet{gibbs2021adaptiveconformalinferencedistribution}).} 
\textit{With probability one, we have that for all $t \in \mathbb{N}$, $\alpha_t \in[-\gamma, 1 + \gamma]$.}

\vspace{1.5mm}
\noindent \textit{Proof of Lemma 4.} 
Assume by contradiction that with positive probability, the sequence $\{\alpha_t\}_{t \in \mathbb{N}}$ is such that $\inf_t \alpha_t < -\gamma$ (the case for $\sup_t \alpha_t > 1 + \gamma$ is symmetric). Notice that the maximum change in one step is bounded: $\sup_t |\alpha_{t+1} - \alpha_t| = \sup_t \gamma|\rho - \text{err}_t| < \gamma$. 
Thus, with positive probability, we may find a specific time step $t \in \mathbb{N}$ such that $\alpha_t < 0$ and $\alpha_{t+1} < \alpha_t$. 

However, by the boundary definition of the quantile function:
{\small
\begin{equation}
    \alpha_t < 0 \implies \hat{Q}_t(1 - \alpha_t) = \infty \implies \text{err}_t = 0
\end{equation}
}
Substituting $\text{err}_t = 0$ into the update rule gives:
\begin{equation}
    \alpha_{t+1} = \alpha_t + \gamma(\alpha - 0) \geq \alpha_t
\end{equation}
This contradicts the assumption that $\alpha_{t+1} < \alpha_t$. Thus, $\mathbb{P}(\exists t \text{ such that } \alpha_{t+1} < \alpha_t < 0) = 0$, establishing the lower bound. 

\vspace{1.5mm}
\noindent \textbf{Theorem 5} \textit{(Distribution-free asymptotic coverage, Proposition 4.1 in \citet{gibbs2021adaptiveconformalinferencedistribution}).} 
\textit{With probability one, for all horizon lengths $T \in \mathbb{N}$, the empirical miscoverage rate satisfies:}
{\small
\begin{equation}
    \label{eq:aci_bound}
    \left| \frac{1}{T} \sum_{t=1}^T \text{err}_t - \rho \right| \leq \frac{\max\{\alpha_1, 1 - \alpha_1\} + \gamma}{T\gamma}
\end{equation}
}
\textit{In particular, as $T \to \infty$, the average miscoverage converges almost surely to the target rate $\alpha$:}
\begin{equation}
    \lim_{T \to \infty} \frac{1}{T} \sum_{t=1}^T \text{err}_t \overset{a.s.}{=} \rho
\end{equation}

\vspace{1.5mm}
\noindent \textit{Proof of Theorem 5.} 
By recursively expanding the update rule $\alpha_{t+1} = \alpha_t + \gamma(\alpha - \text{err}_t)$ from $t=1$ to $T$, we obtain the telescoping sum:
\begin{equation}
    \alpha_{T+1} = \alpha_1 + \sum_{t=1}^T \gamma(\rho - \text{err}_t)
\end{equation}
Rearranging the terms to isolate the empirical average of $\text{err}_t$, we get:
\begin{equation}
    \frac{1}{T} \sum_{t=1}^T (\text{err}_t - \rho) = \frac{\alpha_1 - \alpha_{T+1}}{T\gamma}
\end{equation}
Taking the absolute value on both sides yields:
\begin{equation}
    \left| \frac{1}{T} \sum_{t=1}^T \text{err}_t - \rho \right| = \frac{|\alpha_1 - \alpha_{T+1}|}{T\gamma}
\end{equation}
From Lemma 4, we know that $\alpha_{T+1} \in [-\gamma, 1 + \gamma]$. Given that the initialization $\alpha_1 \in [0, 1]$, the maximum possible distance between $\alpha_1$ and $\alpha_{T+1}$ is bounded by:
{\small
\begin{equation}
    |\alpha_1 - \alpha_{T+1}| \leq \max\{\alpha_1 - (-\gamma), 1 + \gamma - \alpha_1\} 
\end{equation}
}
which simplifies to $\max\{\alpha_1, 1 - \alpha_1\} + \gamma$. Substituting this upper bound into the absolute difference completes the proof for Equation \ref{eq:aci_bound}. 

Taking the limit as $T \to \infty$, the right-hand side of Equation \ref{eq:aci_bound} diminishes to zero (since $\gamma$ is a fixed positive constant), proving that ACI flawlessly achieves the exact marginal coverage frequency over time, without making any assumptions on the nature of the data distribution shift. 

\section{Supplementary Implementation Details}
\label{sec:experimental_details}

\paragraph{Environment}
Our framework operates on a dual-service architecture developed based on the VeRL distributed reinforcement learning framework. The Training Service executes GRPO using Python 3.12, PyTorch 2.8.0 (CUDA 12.9), and is distributed across 4 GPUs via Ray (v2.49.2). To accelerate asynchronous multi-turn rollouts, it leverages sglang (v0.5.3rc0) equipped with the flashinfer backend and flash-attn (v2.8.3). The Retrieval Service operates independently as a FastAPI-based REST endpoint using Python 3.10 and PyTorch 2.4.0 (CUDA 12.1). It utilizes faiss-gpu (v1.8.0) and the e5-base-v2 embedding model, performing high-throughput dense retrieval via mean pooling on 256-token inputs with FP16 precision. The retrieval backend is configured to handle a peak rate of 120 queries per second (QPS) with a 30-second timeout.

\paragraph{Configurations}
This encompasses our data processing, optimization, and CP settings. Data \& Rollout: Models are trained on a unified search-integrated reasoning dataset in Parquet format. We set the maximum prompt, response, and context lengths to 4096, 3000, and 15,000 tokens, respectively, filtering out prompts that exceed the limit. During the GRPO step, we sample $G=5$ trajectories per prompt with a maximum of 4 assistant turns. Optimization: The Actor is optimized with a learning rate of $1 \times 10^{-6}$ and a warmup ratio of 0.285 (100 steps), while the Critic uses $1 \times 10^{-5}$. Training employs a global batch size of 512, a low-variance KL penalty coefficient of 0.001, and Fully Sharded Data Parallel (FSDP) with tensor model parallelism set to 1. Reward Design: The rule-based reward comprises an EM accuracy score (weight 1.0) and format rewards (0.2 for structural integrity, 0.1 for the final answer boundary). CAS: On the retrieval side, APS are applied with a significance level $\alpha_{APS} = 0.20$ and a temperature of 0.01, dynamically restricting the retrieved subset to between 1 and 5 documents. On the training side, ACI is initialized with a target error rate $\rho = 0.25$ and an update step size $\gamma = 0.005$. We apply a discount factor $\eta = 0.5$ to penalize low-confidence trajectories ($s_i > \hat{q}_t$), while empirical error tracking utilizes an Exponential Moving Average (EMA) ratio of 0.05 to maintain quantile stability.

\paragraph{Hardware}
All experiments were conducted on a single server node. The server is configured with dual-socket AMD EPYC 9454 48-Core processors, providing a total of 96 physical cores and 192 threads, organized into two NUMA nodes. The server is equipped with four NVIDIA RTX PRO 6000 Blackwell Server Edition GPUs and 755 GiB of system memory. Storage infrastructure includes a 446.6 GB drive for the OS and environment, alongside a 14.6 TB enterprise-grade drive for high-throughput data caching. The software environment is built on Ubuntu 24.04.3 LTS.

\section{Prompts}
\label{sec:prompts}
In this section, we present the detailed prompt templates utilized across different stages of CAS. The configuration of the reasoning template in Table~\ref{tab:prompt_reasoning} is adapted from Search-R1 \cite{jin2025searchr1trainingllmsreason} to maintain consistency in agentic behavior. Additionally, the query decomposition prompt in Table~\ref{tab:prompt_decomposition} and the retrieval relevance judge prompt in Table~\ref{tab:prompt_judge} are specifically employed to construct the calibration set for the retrieval-side APS.

\section{Additional Experimental Analysis on Qwen3-8B}

\label{sec:appendix_dynamics}
\begin{figure*}[htbp]
    \centering
    \begin{subfigure}[b]{0.24\textwidth}
        \centering
        \includegraphics[width=\textwidth]{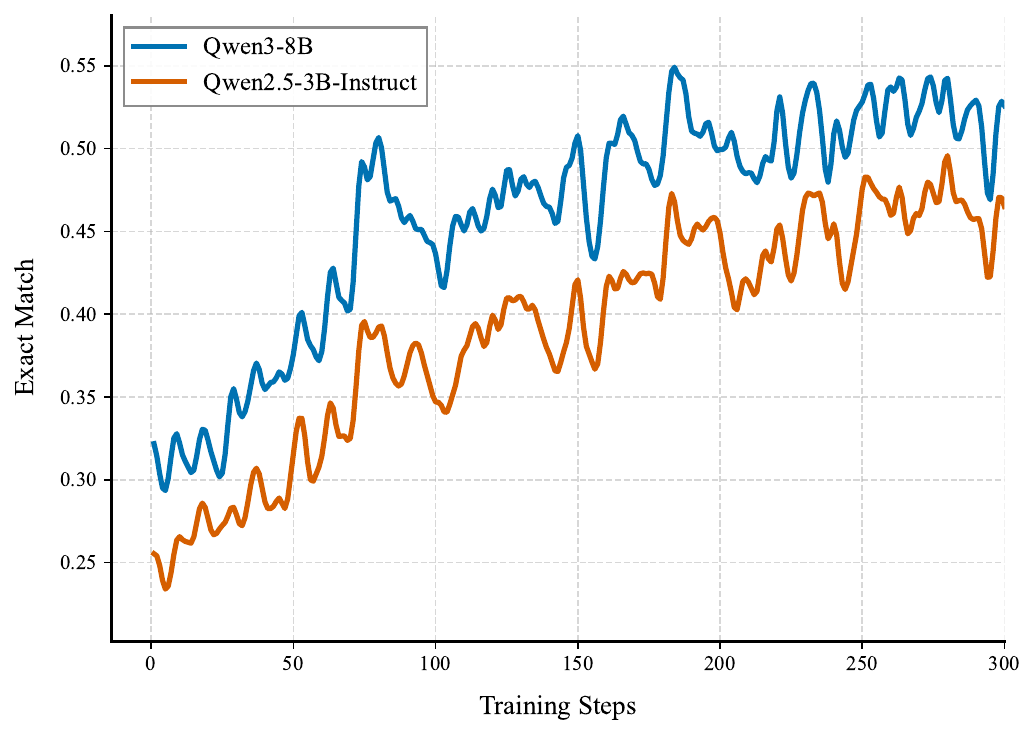}
        \caption{EM score during training.}
        \label{fig:em_train}
    \end{subfigure}
    \hfill
    \begin{subfigure}[b]{0.24\textwidth}
        \centering
        \includegraphics[width=\textwidth]{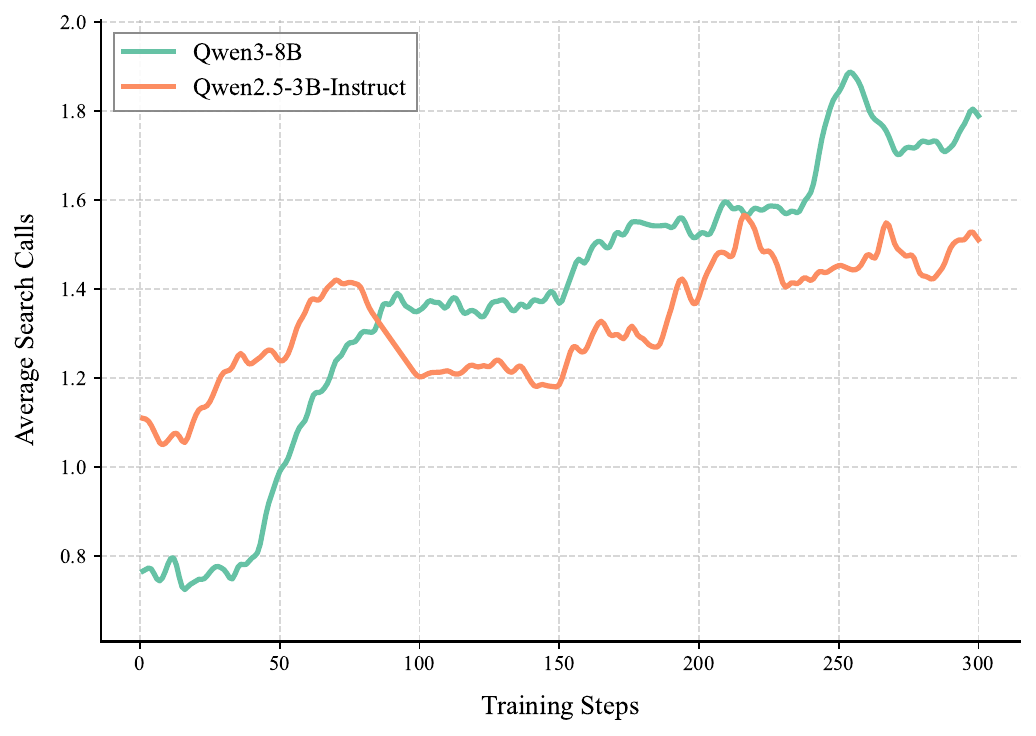}
        \caption{Average search calls.}
        \label{fig:search_calls}
    \end{subfigure}
    \hfill
    \begin{subfigure}[b]{0.24\textwidth}
        \centering
        \includegraphics[width=\textwidth]{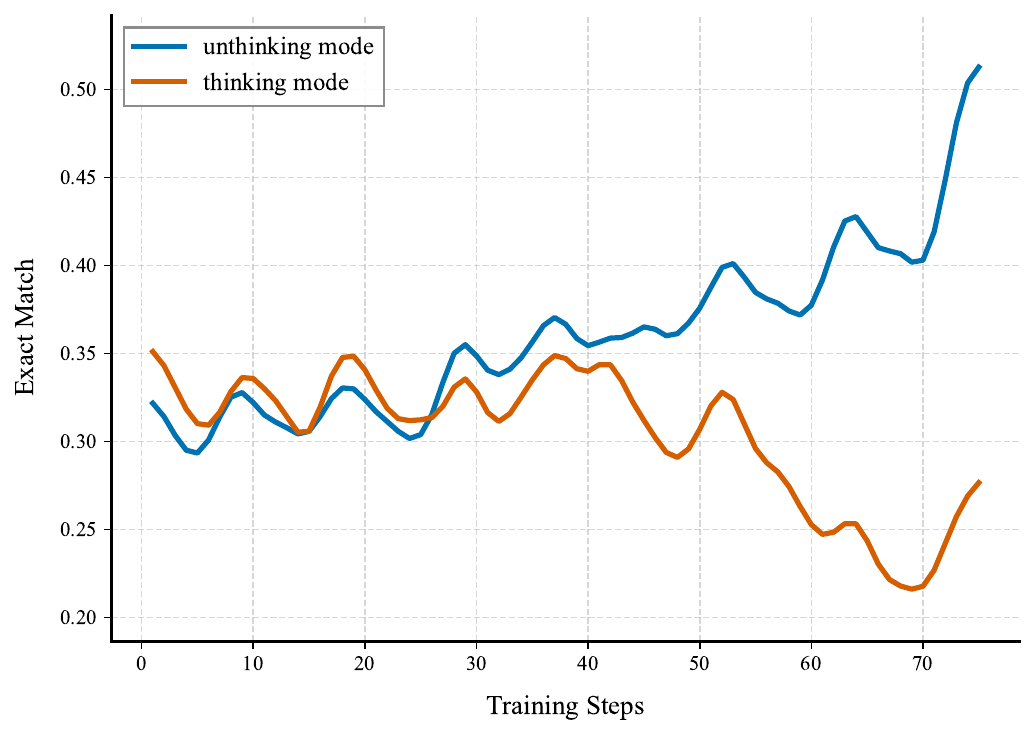}
        \caption{EM (Thinking mode).}
        \label{fig:em_thinking}
    \end{subfigure}
    \hfill
    \begin{subfigure}[b]{0.24\textwidth}
        \centering
        \includegraphics[width=\textwidth]{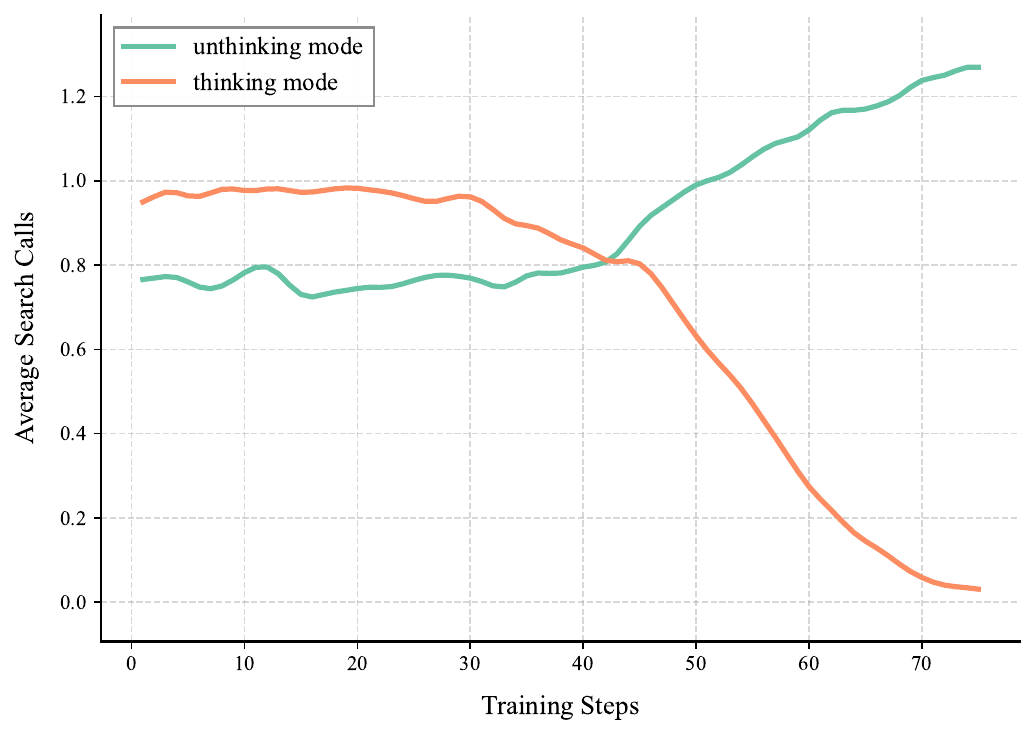}
        \caption{Calls (Thinking mode).}
        \label{fig:search_thinking}
    \end{subfigure}

    \caption{Visualization of training metrics and the impact of the thinking mode. Panels (a) and (b) illustrate the progression of accuracy and tool invocation frequency during training. Panels (c) and (d) demonstrate how the thinking mode influences the final EM scores and the efficiency of the search process.}
    \label{fig:training_dynamics}
\end{figure*}

In Table~\ref{tab:main_results}, it is observed that our method's performance on Qwen3-8B consistently outperforms that on Qwen2.5-3B-Instruct. This improvement is primarily attributed to the inherent model capacity of Qwen3-8B, which exhibits a significant advantage over the 3B-Instruct variant, as illustrated in Figure~\ref{fig:em_train}. Regarding the search behavior shown in Figure~\ref{fig:search_calls}, we note that the average search calls for Qwen3-8B remain lower than those of Qwen2.5-3B-Instruct during approximately the first 80 training steps. This phenomenon occurs because Qwen2.5-3B-Instruct, as a smaller model, tends to exhibit erratic and indiscriminate tool invocation during the early stages of training. In contrast, the larger parameter scale of Qwen3-8B ensures more efficient search calls from the beginning. This efficiency is further evidenced by comparing Figure~\ref{fig:em_train} and Figure~\ref{fig:search_calls}, where Qwen3-8B achieves substantially higher EM scores despite a noticeably lower frequency of search invocations. Furthermore, as depicted in Figure~\ref{fig:em_train}, although the number of search calls for Qwen3-8B increases slightly relative to Qwen2.5-3B-Instruct after convergence, it remains significantly more efficient than baseline methods lacking ACI constraints. This demonstrates that the ACI mechanism effectively modulates low-confidence trajectories even when applied to the 8B model.

For the experiments involving Qwen3-8B, the thinking mode is disabled by default, as enabling this feature leads to a drastic reduction in training effectiveness, as shown in Figure~\ref{fig:em_thinking}. The underlying cause is revealed in Figure~\ref{fig:search_thinking}: after enabling the thinking mode, the model initially tends towards aggressive search calls due to the interleaving of internal reasoning with our prescribed reasoning grammar. However, the model rapidly discovers that many single-hop problems can be resolved solely through internal reasoning. Consequently, it gradually ceases to invoke the search tool, leading to a complete cessation of active information gathering and rendering the training process ineffective for the intended search-integrated tasks.


\begin{table*}[t]
\centering
\small
\begin{tabular}{p{0.95\textwidth}}
\toprule
\textbf{Prompt 1: Template for CAS} \\
\midrule
You are Qwen, created by Alibaba Cloud. You are a helpful assistant. Answer the given question. You must conduct reasoning inside \texttt{<think>} and \texttt{</think>} first every time you get new information. After reasoning, if you find you lack some knowledge, you can call a search engine by \texttt{<search>query</search>} and it will return the top searched results between \texttt{<information>} and \texttt{</information>}. You can search as many times as you want. If you find no further external knowledge needed, you can directly provide the answer inside \texttt{<answer>} and \texttt{</answer>}, without detailed illustrations. For example, \texttt{<answer> Beijing </answer>}. \\
\\
Question: [Question] \\
\bottomrule
\end{tabular}
\caption{Template for CAS reasoning process.}
\label{tab:prompt_reasoning}
\end{table*}

\begin{table*}[t]
\centering
\small
\begin{tabular}{p{0.95\textwidth}}
\toprule
\textbf{Prompt 2: Query Decomposition} \\
\midrule
You decompose QA tasks into hop-level search query-answer pairs. Return strict JSON. Given a QA sample, split it into single-hop query-answer pairs. \\
\\
Rules: \\
1) If single-hop, return one pair. \\
2) If multi-hop, return one pair per hop. \\
3) answer should be concise and factual. \\
4) Output JSON only in schema: \texttt{\{"pairs":[\{"query":"...","answer":"..."\}]\}} \\
\\
question: \{question\} \\
ground\_truth\_answers: \{gt\_answers\} \\
metadata: \{metadata\} \\
extra\_info: \{extra\_info\} \\
\bottomrule
\end{tabular}
\caption{Prompt for decomposing multi-hop queries into single-hop sub-queries.}
\label{tab:prompt_decomposition}
\end{table*}

\begin{table*}[t]
\centering
\small
\begin{tabular}{p{0.95\textwidth}}
\toprule
\textbf{Prompt 3: Retrieval Relevance Judge} \\
\midrule
You are a precise retrieval relevance judge for QA reasoning. You are given a retrieval query, its target answer, and retrieved documents. Decide which document(s) can support reasoning to the target answer. \\
\\
Return strict JSON with schema: \\
\texttt{\{"golden\_doc\_indices":[0,1], "best\_golden\_doc\_index":0, "reason":"..."\}} \\
\\
Rules: \\
1) If none can support the answer, return empty golden\_doc\_indices and -1 as best index. \\
2) best\_golden\_doc\_index must be one item in golden\_doc\_indices, or -1. \\
3) Never output markdown. \\
\\
query: \{query\} \\
answer: \{answer\} \\
retrieved\_docs: \{docs\_for\_llm\} \\
\bottomrule
\end{tabular}
\caption{Judge prompt for locating the most relevant documents.}
\label{tab:prompt_judge}
\end{table*}

\section{Detailed Results for Ablation and Sensitivity Analysis}
\label{sec:appendix_ablation_sensitivity}
Corresponding to Table~\ref{tab:ablation} in the main text, we provide the complete results across all individual datasets in Table~\ref{tab:detailed_ablation}.

\begin{table*}[t]
\centering
\caption{Detailed ablation study and sensitivity analysis on Qwen2.5-3B-Instruct. The table presents component-wise ablations and sensitivity analyses for $\rho$ and $\alpha_{\text{APS}}$.}
\label{tab:detailed_ablation}
\resizebox{\textwidth}{!}{
\begin{tabular}{lcccccccc}
\toprule
\multirow{2}{*}{Methods} & \multicolumn{3}{c}{General QA} & \multicolumn{4}{c}{Multi-Hop QA} & \multirow{2}{*}{Average} \\
\cmidrule(lr){2-4} \cmidrule(lr){5-8}
& NQ$^\dagger$ & TriviaQA$^\star$ & PopQA$^\star$ & HotpotQA$^\dagger$ & 2WikiMultiHopQA$^\star$ & Musique$^\star$ & Bamboogle$^\star$ & \\
\midrule
- ACI & 0.417 & 0.598 & 0.430 & 0.378 & 0.372 & 0.143 & 0.352 & 0.384 \\
- APS & 0.435 & 0.603 & 0.433 & 0.393 & 0.374 & 0.143 & 0.344 & 0.389 \\
\midrule
$\rho = 0.1$ & 0.441 & 0.601 & 0.449 & 0.382 & 0.375 & 0.132 & 0.320 & 0.386 \\
$\rho = 0.4$ & 0.438 & 0.596 & 0.452 & 0.356 & 0.351 & 0.113 & 0.272 & 0.368 \\
\midrule
$\alpha_{\text{APS}} = 0.35$ & 0.420 & 0.584 & 0.432 & 0.372 & 0.370 & 0.146 & 0.288 & 0.372 \\
$\alpha_{\text{APS}} = 0.05$ & 0.442 & 0.613 & 0.459 & 0.328 & 0.285 & 0.076 & 0.248 & 0.350 \\
\bottomrule
\end{tabular}
}
\end{table*}

\section{Case Study}
\label{sec:appendix_case_study}
In this section, we present representative qualitative cases to illustrate the core behavioral patterns of CAS.

Table~\ref{tab:case_simple_single_hop} illustrates a straightforward single-hop scenario. The model directly addresses the factual query by formulating a precise search action. Because the retrieved documents are clean and highly relevant, the agent swiftly concludes its reasoning and extracts the correct answer without unnecessary actions.

Table~\ref{tab:case_complex_single_hop} presents a more challenging single-hop case characterized by high retrieval noise. Although the search results contain highly distracting entities with similar names, the agent successfully evaluates the contextual relevance of each document, filters out the irrelevant distractors, and accurately grounds its final answer on the correct source.

Table~\ref{tab:case_multi_hop} demonstrates the framework's capability to handle multi-hop queries through interleaved reasoning and search. The agent dynamically decomposes the complex task in its initial \texttt{<think>} block, retrieves the missing bridge entity in the first hop, and uses this intermediate information to construct a targeted query for the subsequent hop. This iterative process highlights the effectiveness of allowing the model to flexibly transition between internal deliberation and external tool interaction.

\begin{table*}[t]
\centering
\small
\begin{tabular}{p{0.96\textwidth}}
\toprule
\textit{Question:} What BBC television drama serial starring Jim McManus was based on Sarah Waters' debut novel of the same name? \\
\textit{Golden Answer:} ['Tipping the Velvet'] \\
\midrule
\texttt{<think>} \\
The question asks for a BBC television drama serial based on Sarah Waters' debut novel of the same name. I recall that Sarah Waters' debut novel is probably "Tipping the Velvet", but I need to confirm it, especially because Jim McManus may create ambiguity in search results. \\
\texttt{</think>} \\
\\
\texttt{<search>} Jim McManus Tipping the Velvet \texttt{</search>} \\
\\
\texttt{<information>} \\
Doc 1 (Title: "Jack McManus") \\
Jack McManus Jack McManus (born Thomas McManus, March 1862 – May 1905), also known as Eat 'Em Up, was a noted New York City gangster around the turn of the 20th century. Born in Boston, he was considered one of the premier boxers of the underworld, rivaled only by Monk Eastman. \\
\\
Doc 2 (Title: "Jimmy Velvet") \\
toured in 1973-1978 with the Jimmy Velvet Show band. Band members included Steve Morgan (guitar), John Collins (bassist), Dennis Moog (drums), and Kathy Slinkard Velvet. Velvet was championed by TV host Dick Clark, who had him on his "American Bandstand" show more than once. \\
\\
Doc 3 (Title: "Tipping the Velvet") \\
Tipping the Velvet is a historical novel published as Sarah Waters' debut novel in 1998. Set in England during the 1890s, it tells a coming of age story about a young woman named Nan who falls in love with a male impersonator, follows her to London, and finds various ways to support herself as she journeys through the city. \\
\\
Doc 4 (Title: "James McManus") \\
James "Jim" McManus (born March 22, 1951) is an American teacher, writer and poker player living in Kenilworth, Illinois. McManus is best known as the author of the book "Positively Fifth". \\
\texttt{</information>} \\
\\
\texttt{<think>} \\
The search results contain several noisy results about other people named McManus or Velvet, but Doc 3 directly confirms that "Tipping the Velvet" is Sarah Waters' debut novel. Since the question asks for the BBC drama serial based on that debut novel of the same name, the answer is "Tipping the Velvet". \\
\texttt{</think>} \\
\\
\texttt{<answer>} Tipping the Velvet \texttt{</answer>} \\
\bottomrule
\end{tabular}
\caption{Case Study: Complex Single-Hop QA.}
\label{tab:case_complex_single_hop}
\end{table*}

\begin{table*}[t]
\centering
\small
\begin{tabular}{p{0.96\textwidth}}
\toprule
\textit{Question:} Who won the Men's Singles French Open Tennis Tournament in 2018? \\
\textit{Golden Answer:} ['Rafael Nadal'] \\
\midrule
\texttt{<think>} \\
I need to find out who won the Men's Singles French Open Tennis Tournament in 2018. I remember Rafael Nadal was dominant at the French Open around that time, but I should verify the exact 2018 result. \\
\texttt{</think>} \\
\\
\texttt{<search>} 2018 French Open Men's Singles winner \texttt{</search>} \\
\\
\texttt{<information>} \\
Doc 1 (Title: "2018 French Open – Men's Singles") \\
2018 French Open – Men's Singles Rafael Nadal was the defending champion and successfully defended his title, defeating Dominic Thiem in the final, 6–4, 6–3, 6–2. Nadal equaled Margaret Court's all-time record of 11 Grand Slam singles titles won at one major and became the first player to achieve that feat in the Open era. He dropped only one set during the entire tournament (losing the first set of his quarterfinal match against Diego Schwartzman). Nadal also retained the ATP No.\ 1 singles ranking by defending the title. \\
\\
Doc 2 (Title: "2018 French Open – Men's Singles") \\
72nd in the ATP rankings, became the first unseeded men's singles semifinalist at the French Open since Gaël Monfils in 2008, the lowest-ranked male singles player to reach the French Open semifinals since Andriy Medvedev in 1999, and the first Italian man to reach a Grand Slam singles semifinal since Corrado Barazzutti at the 1978 French Open. 2018 French Open – Men's Singles Rafael Nadal was the defending champion and successfully defended his title, defeating Dominic Thiem in the final, 6–4, 6–3, 6–2. \\
\texttt{</information>} \\
\\
\texttt{<think>} \\
The search results clearly state that Rafael Nadal won the 2018 French Open men's singles title by defeating Dominic Thiem in the final. \\
\texttt{</think>} \\
\\
\texttt{<answer>} Rafael Nadal \texttt{</answer>} \\
\bottomrule
\end{tabular}
\caption{Case Study: Simple Single-Hop QA.}
\label{tab:case_simple_single_hop}
\end{table*}

\begin{table*}[t]
\centering
\small
\begin{tabular}{p{0.96\textwidth}}
\toprule
\textit{Question:} Which hockey team does the younger brother of Sean James Pronger currently work for? \\
\textit{Golden Answer:} ['the Florida Panthers'] \\
\midrule
\texttt{<think>} \\
I need to identify Sean James Pronger's younger brother first, then determine which hockey team that person currently works for. This requires at least two steps: family relation lookup, then current team lookup. \\
\texttt{</think>} \\
\\
\texttt{<search>} Sean James Pronger younger brother \texttt{</search>} \\
\\
\texttt{<information>} \\
Doc 1 (Title: "Jordan Reaves") \\
with Saskatchewan for 2018 training camp. Jordan is the youngest son of former Winnipeg Blue Bomber running back Willard Reaves, who played with the team for five years and was a member of the 72nd Grey Cup winning team. Jordan is also the younger brother of Ryan Reaves, a professional hockey player and right wing for the Vegas Golden Knights of the National Hockey League. \\
\\
Doc 2 (Title: "Chris Pronger") \\
the 1990–91 season. One of his defence partners in Stratford was future NHL player Greg de Vries. In May 1991, Pronger indicated he was going to join his older brother Sean at Bowling Green State University (NCAA) instead of opting for the OHL. Regardless of his pre-draft indications, Pronger was selected in the 6th round by the Peterborough Petes in the OHL Priority Selection. \\
\\
Doc 3 (Title: "Luke Schenn") \\
hits. He believes he is at his best when he is playing a tough game, while staying responsible. Growing up, Schenn patterned his play after fellow defencemen Chris Pronger and Rob Blake. Schenn was born on November 2, 1989, in Saskatoon, Saskatchewan. His younger brother Brayden was picked fifth overall in the 2009 NHL Entry Draft by the Los Angeles Kings. \\
\\
Doc 4 (Title: "Jordan Eberle") \\
two sisters, Ashley and Whitney, and a younger brother, Dustin. He attended high school at the Athol Murray College of Notre Dame in Wilcox, Saskatchewan. His brother Dustin was drafted. \\
\texttt{</information>} \\
\\
\texttt{<think>} \\
Doc 2 says Chris Pronger planned to join his older brother Sean at Bowling Green State University. Therefore, Sean James Pronger's younger brother is Chris Pronger. Now I need to find which hockey team Chris Pronger currently works for. \\
\texttt{</think>} \\
\\
\texttt{<search>} Chris Pronger current team \texttt{</search>} \\
\\
\texttt{<information>} \\
Doc 1 (Title: "Chris Pronger") \\
Chris Pronger Christopher Robert Pronger (born October 10, 1974) is a Canadian former professional ice hockey defenceman who is currently the senior advisor of hockey operations for the Florida Panthers. He had not played since November 2011 due to post-concussion syndrome. \\
\\
Doc 2 (Title: "Chris Pronger") \\
Chris Pronger Christopher Robert Pronger is a Canadian former professional ice hockey defenceman who is currently the senior advisor of hockey operations for the Florida Panthers. \\
\texttt{</information>} \\
\\
\texttt{<think>} \\
The second search confirms that Chris Pronger currently works as senior advisor of hockey operations for the Florida Panthers. Therefore, the hockey team is the Florida Panthers. \\
\texttt{</think>} \\
\\
\texttt{<answer>} Florida Panthers \texttt{</answer>} \\
\bottomrule
\end{tabular}
\caption{Case Study: Multi-Hop QA.}
\label{tab:case_multi_hop}
\end{table*}

\end{document}